\documentstyle[aclnobib]{article}

\author{Mark Lauer \\
Microsoft Institute \\
65 Epping Road, \\
North Ryde NSW 2113 \\
Australia \\{\tt t-markl@microsoft.com}
}

\title{ Corpus Statistics Meet the Noun Compound: \\
	Some Empirical Results }

\newcommand{\nn}{$_{\mbox{\footnotesize\sc n}}$}

\newtheorem{example}{Example}

\begin{document}


\maketitle

\begin{abstract}
A variety of statistical methods for noun compound analysis
are implemented and compared.  The results 
support two main conclusions.  First, the use of conceptual 
association not only enables a broad coverage, but
also improves the accuracy.  Second, an analysis
model based on dependency grammar is substantially
more accurate than one based on deepest constituents,
even though the latter is more prevalent in the
literature.
\end{abstract}

\section{Background}

\subsection{Compound Nouns}
If parsing is taken to be the first step in taming the 
natural language understanding task, then broad coverage 
NLP remains a jungle inhabited by wild beasts.
For instance, parsing noun compounds appears to 
require detailed world knowledge that is unavailable
outside a limited domain (Sparck Jones, 1983).  
Yet, far from being an obscure, endangered species, 
the noun compound is flourishing in modern language.  
It has already made five appearances in this paragraph 
and at least one diachronic study shows a veritable population
explosion (Leonard, 1984).  While substantial work on noun
compounds exists in both linguistics (e.g. Levi, 1978; Ryder, 1994) 
and computational linguistics (Finin, 1980; McDonald, 
1982; Isabelle, 1984), techniques suitable for broad 
coverage parsing remain unavailable.  This paper explores
the application of corpus statistics (Charniak, 1993)
to noun compound parsing (other computational problems are
addressed in Arens {\em et~al}, 1987; Vanderwende, 1993 
and Sproat, 1994).

The task is illustrated in example \ref{two_ambig_compounds}:
\begin{example} \label{two_ambig_compounds} 
\hspace{1in} 
\end{example}
\begin{tabbing}
(a) [woman\nn\ [aid\nn\ worker\nn]] \\
(b) [[hydrogen\nn\ ion\nn] exchange\nn]   
\end{tabbing}
The parses assigned to these two compounds differ,
even though the sequence of parts of speech are
identical.  The problem is analogous to the 
prepositional phrase attachment task explored in Hindle and
Rooth (1993).  The approach they propose involves computing
lexical associations from a corpus and using these to select the
correct parse.  A similar architecture may be applied to noun
compounds.  

In the experiments below the accuracy of such a 
system is measured.  Comparisons are made across five dimensions:
\begin{itemize}
\item Each of two analysis models are applied: adjacency and dependency.
\item Each of a range of training schemes are employed.
\item Results are computed with and without tuning factors suggested in the literature.
\item Each of two parameterisations are used: associations 
 between words and associations between concepts.
\item Results are collected with and without machine tagging 
 of the corpus.
\end{itemize}

\subsection{Training Schemes}
While Hindle and Rooth (1993) use a partial parser to 
acquire training data, such machinery appears 
unnecessary for noun compounds.  Brent (1993)
has proposed the use of simple word patterns for the
acquisition of verb subcategorisation information.
An analogous approach to compounds is used in
Lauer (1994) and constitutes one scheme evaluated
below.  While such patterns produce false
training examples, the resulting noise often only
introduces minor distortions.

A more liberal alternative is the use of a 
co-occurrence window.  Yarowsky (1992)
uses a fixed 100 word window to collect 
information used for sense disambiguation.  
Similarly, Smadja (1993) uses a six
content word window to extract significant
collocations.  A range of windowed training
schemes are employed below.  Importantly, 
the use of a window provides a natural
means of trading off the amount of data against its
quality.  When data sparseness undermines the
system accuracy, a wider window may admit a
sufficient volume of extra accurate data to outweigh
the additional noise.  

\subsection{Noun Compound Analysis} \label{intro_analysis}
There are at least four existing corpus-based
algorithms proposed for syntactically analysing
noun compounds.  Only two of these have
been subjected to evaluation, and in each case, no
comparison to any of the other three was performed.
In fact all authors appear unaware of the other three
proposals.  I will therefore briefly describe these
algorithms.

Three of the algorithms use what I will call the {\sc
adjacency model}, an analysis procedure that goes
back to Marcus (1980, p253).  Therein, the
procedure is stated in terms of calls to an oracle
which can determine if a noun compound is
acceptable.  It is reproduced here for reference:

Given three nouns $n_1$, $n_2$ and $n_3$:
\begin{itemize}
\item If either [$n_1$ $n_2$] or [$n_2$ $n_3$] is not
semantically acceptable then build the alternative structure;
\item otherwise, if [$n_2$ $n_3$] is semantically
preferable to [$n_1$ $n_2$] then build [$n_2$ $n_3$];
\item otherwise, build [$n_1$ $n_2$].
\end{itemize}

Only more recently has it been suggested that corpus
statistics might provide the oracle, and this idea is
the basis of the three algorithms which use the
adjacency model.  The simplest of these is
reported in Pustejovsky {\em et al}  (1993).  Given a three
word compound, a search is conducted elsewhere in
the corpus for each of the two possible
subcomponents.  Whichever is found is then chosen
as the more closely bracketed pair.  For example,
when {\em backup compiler disk} is encountered,
the analysis will be:
\begin{example} \label{adj_example}
\hspace{1in}
\end{example}
\begin{tabbing}
(a) [backup\nn\ [compiler\nn\ disk\nn]] \\
\hspace{0.4in} \= when {\em compiler disk} appears elsewhere \\
(b) [[backup\nn\ compiler\nn] disk\nn] \\
	       \> when {\em backup compiler} appears elsewhere
\end{tabbing}
Since this is proposed merely as a rough heuristic, it
is not stated what the outcome is to be if neither or
both subcomponents appear.  Nor is there any
evaluation of the algorithm.

The proposal of Liberman and Sproat (1992) is
more sophisticated and allows for the frequency of
the words in the compound.  Their proposal
involves comparing the mutual information between
the two pairs of adjacent words and bracketing
together whichever pair exhibits the highest.  Again, there is no
evaluation of the method other than a demonstration
that four examples work correctly.  

The third proposal based on the adjacency model
appears in Resnik (1993) and is rather more
complex again. The {\sc selectional association}
between a predicate and a word is defined based on
the contribution of the word to the conditional
entropy of the predicate.  The association between
each pair of words in the compound is then
computed by taking the maximum selectional
association from all possible ways of regarding the
pair as predicate and argument.  Whilst this
association metric is complicated, the
decision procedure still follows the outline devised
by Marcus (1980) above.  Resnik (1993) used 
unambiguous noun compounds from the parsed 
{\em Wall Street Journal} (WSJ) corpus to estimate 
the association values and analysed a test set of 
around 160 compounds.  After some tuning, the accuracy
was about 73\%, as compared with a baseline of
64\% achieved by always bracketing the first two
nouns together.

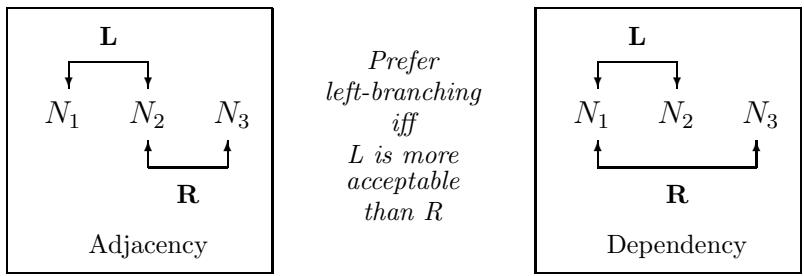
\begin{figure*}
\centering
\begin{picture}(300,100)
\put(200,0){
\begin{picture}(100,100)
\put(0,0){\framebox(100,100){}}
\put(50,60){
   \makebox(0,0){
      \large $N_1$ \hspace{10pt} 
      \large $N_2$ \hspace{10pt} 
      \large $N_3$
   }
}
\put(50,10){
   \makebox(0,0){Dependency}
}
\put(15, 70){
   \begin{picture}(40, 30)(-5,0)
     \put(15,20){\makebox(0,0){\bf L}}
     \put(0,10){\vector(0,-1){10}}
     \put(30,10){\vector(0,-1){10}}
     \put(0,10){\line(1,0){30}}
   \end{picture}
}
\put(15, 20){
   \begin{picture}(70, 30)(-5,0)
      \put(30,10){\makebox(0,0){\bf R}}
      \put(0,20){\vector(0,1){10}}
      \put(60,20){\vector(0,1){10}}
      \put(0,20){\line(1,0){60}}
   \end{picture}
}
\end{picture}
}

\put(120, 20){
   \shortstack{\em Prefer \\ \em left-branching \\ \em iff \\
       \em L is more \\ \em acceptable \\ \em than R }
}

\put(0,0){
\begin{picture}(100,100)
\put(0,0){\framebox(100,100){}}
\put(50,60){
   \makebox(0,0){
      \large $N_1$ \hspace{10pt} 
      \large $N_2$ \hspace{10pt} 
      \large $N_3$
   }
}
\put(50,10){
   \makebox(0,0){Adjacency}
}
\put(15, 70){
   \begin{picture}(40, 30)(-5,0)
     \put(15,20){\makebox(0,0){\bf L}}
     \put(0,10){\vector(0,-1){10}}
     \put(30,10){\vector(0,-1){10}}
     \put(0,10){\line(1,0){30}}
   \end{picture}
}
\put(45, 20){
   \begin{picture}(40, 30)(-5,0)
      \put(15,10){\makebox(0,0){\bf R}}
      \put(0,20){\vector(0,1){10}}
      \put(30,20){\vector(0,1){10}}
      \put(0,20){\line(1,0){30}}
   \end{picture}
}
\end{picture}
}
\end{picture}

\caption{Two analysis models and the associations they compare}
\label{graphical_models}
\end{figure*}

The fourth algorithm, first described in Lauer
(1994), differs in one striking manner from the other
three.  It uses what I will call the {\sc dependency
model}.  This model utilises the following
procedure when given three nouns $n_1$, $n_2$ and $n_3$:
\begin{itemize}
\item Determine how acceptable the structures
[$n_1$ $n_2$] and [$n_1$ $n_3$] are;
\item if the latter is more acceptable, build [$n_2$ $n_3$] first;
\item  otherwise, build [$n_1$ $n_2$] first.
\end{itemize}
Figure \ref{graphical_models} shows a graphical comparison
of the two analysis models.
 
In Lauer (1994), the degree of acceptability is again
provided by statistical measures over a corpus.  The
metric used is a mutual information-like measure
based on probabilities of modification relationships.
This is derived from the idea that parse trees capture
the structure of semantic relationships within a noun
compound.\footnote{Lauer and Dras (1994) give a
formal construction motivating the algorithm given in
Lauer (1994).}

The dependency model attempts to
choose a parse which makes the resulting
relationships as acceptable as possible.  For
example, when {\em backup compiler disk} is
encountered, the analysis will be:
\begin{example} \label{dep_example}
\hspace{1in}
\end{example}
\begin{tabbing}
(a) [backup\nn\ [compiler\nn\ disk\nn]] \\
\hspace{0.4in}  \= when {\em backup disk} is more acceptable \\
(b) [[backup\nn\ compiler\nn] disk\nn] \\
		\> when {\em backup compiler} is more acceptable
\end{tabbing}
I claim that the dependency model makes more
intuitive sense for the following reason.  Consider
the compound {\em calcium ion exchange}, which is
typically left-branching (that is, the first two words
are bracketed together).  There does not seem to be
any reason why {\em calcium ion} should be any
more frequent than {\em ion exchange}.  Both are
plausible compounds and regardless of the
bracketing, {\em ions} are the object of an {\em
exchange}.  Instead, the correct parse depends on
whether {\em calcium} characterises the {\em
ions} or mediates the {\em exchange}.

Another significant difference between the models is
the predictions they make about the proportion of
left and right-branching compounds.  
Lauer and Dras (1994) show that under a dependency
model, left-branching compounds should occur twice as
often as right-branching compounds (that is two-thirds
of the time).  
In the test set used here and in that of Resnik (1993),
the proportion of left-branching compounds is 67\%
and 64\% respectively.  
In contrast, the adjacency model appears to predict 
a proportion of 50\%.

The dependency model has also been 
proposed by Kobayasi {\it et~al} (1994)
for analysing Japanese noun compounds, apparently independently.
Using a corpus to acquire associations, they bracket sequences
of Kanji with lengths four to six (roughly equivalent to two or three
words).  A simple calculation shows that 
using their own preprocessing hueristics to guess a bracketing
provides a higher accuracy on their test set than their statistical
model does.  This renders their experiment inconclusive.

\section{Method}

\subsection{Extracting a Test Set}
A test set of syntactically ambiguous noun
compounds was extracted from our 8 million word
Grolier's encyclopedia corpus in the following 
way.\footnote{We would like to thank Grolier's for 
permission to use this material for research purposes.} 
Because the corpus is not tagged or parsed, a
somewhat conservative strategy of looking for
unambiguous sequences of nouns was used.  To
distinguish nouns from other words, the University
of Pennsylvania morphological analyser (described in
Karp {\it et~al}, 1992)
was used to generate the set of
words that can only be used as nouns (I shall
henceforth call this set $\cal N$).  All consecutive
sequences of these words were extracted, 
and the three word sequences used to form the test set.  
For reasons made clear below, only sequences consisting
entirely of words from Roget's thesaurus
were retained, giving a total of 308 test 
triples.\footnote{The 1911 version of Roget's used is
available on-line and is in the public domain.}

These triples were manually analysed using as context
the entire article in which they appeared.  In some
cases, the sequence was not a noun compound
(nouns can appear adjacent to one another across
various constituent boundaries) and was marked as
an error.  Other compounds exhibited what Hindle
and Rooth (1993) have termed {\sc semantic
indeterminacy} where the two possible bracketings
cannot be distinguished in the context.  The
remaining compounds were assigned either a 
left-branching or right-branching analysis.  Table
\ref{test_dist} shows the number of each kind and
an example of each.

\begin{table*}
\centering
\begin{tabular}{|l|r|r|l|} \hline
Type  & Number & Proportion & Example \\ \hline

Error & 29 & 9\% & In {\em monsoon regions
rainfall} does not \ldots \\

Indeterminate & 35 & 11\% & Most advanced aircraft have 
				{\em precision navigation systems}. \\

Left-branching & 163 & 53\% & \ldots escaped punishment by 
				the Allied {\em war crimes tribunals}. \\

Right-branching & 81 & 26\% & Ronald Reagan, who won two {\em
			landslide election victories}, \ldots \\ \hline
\end{tabular}
\caption{Test set distribution}  \label{test_dist}
\end{table*}

Accuracy figures in all the results reported below
were computed using only those 244 compounds
which received a parse.

\subsection{Conceptual Association}
One problem with applying lexical association to
noun compounds is the enormous number of
parameters required, one for every possible pair of
nouns.
Not only does this
require a vast amount of memory space, it creates a
severe data sparseness problem since we require at
least some data about each parameter.  Resnik and
Hearst (1993) coined the term {\sc conceptual
association} to refer to association values computed
between groups of words.  By assuming that all
words within a group behave similarly, the
parameter space can be built in terms of the
groups rather than in terms of the words.

In this study, conceptual association is used with groups 
consisting of all categories from the 1911 version 
of Roget's thesaurus.\footnote{It contains 1043 categories.}
Given two thesaurus categories $t_1$ and $t_2$, 
there is a parameter which represents the degree of
acceptability of the structure $[n_1 n_2]$ where
$n_1$ is a noun appearing in $t_1$ and $n_2$
appears in $t_2$.  By the assumption that words
within a group behave similarly, this is constant
given the two categories.  Following Lauer and Dras
(1994) we can formally write this parameter as $Pr(
t_1 \rightarrow t_2)$
where the event $t_1 \rightarrow t_2$ denotes the
modification of a noun in $t_2$ by a noun in $t_1$.

\subsection{Training} \label{method_train}
To ensure that the test set is disjoint from the
training data, all occurrences of the test noun
compounds have been removed from the training
corpus.  Two types of training scheme are explored in this
study, both unsupervised.  The first employs a pattern 
that follows Pustejovsky (1993) in counting the
occurrences of subcomponents.
A training instance is any sequence of four words 
$w_1 w_2 w_3 w_4$ where $w_1, w_4 \notin {\cal N}$ 
and $w_2, w_3 \in {\cal N}$.
Let $\mbox{count}_p(n_1, n_2)$ be the
number of times a sequence $w_1 n_1 n_2 w_4$
occurs in the training corpus with $w_1, w_4 \notin
{\cal N}$.

The second type uses a window to collect training
instances by observing how often a pair of nouns 
co-occur within some fixed number of words.  
In this study, a variety of window sizes are used.
For $n \geq 2$, let $\mbox{count}_n(n_1, n_2)$ be the number 
of times a sequence $n_1 w_1 \ldots w_i n_2$ occurs in the
training corpus where $i \leq n-2$.  
Note that windowed counts are asymmetric.
In the case of a window two words wide,
this yields the mutual information metric proposed by
Liberman and Sproat (1992).

Using each of these different training schemes to
arrive at appropriate counts it is then possible to
estimate the parameters.  Since these are
expressed in terms of categories rather than
words, it is necessary to combine the counts of
words to arrive at estimates.  In
all cases the estimates used are:

\begin{displaymath}
\Pr(t_1 \rightarrow t_2) =
\frac{1}{\eta}
\sum_{
\begin{array}{c}
\scriptstyle w_1 \scriptstyle \in \scriptstyle t_1 \\
\scriptstyle w_2 \scriptstyle \in \scriptstyle t_2
\end{array}
}
\frac
{\mbox{count($w_1, w_2\!$)}}
{\mbox{ambig($w_1\!$)} \mbox{ambig($w_2\!$)}}
\end{displaymath}
where
\begin{math}
\eta =
\sum_{
\begin{array}{c}
\scriptstyle w_1 \scriptstyle \in \scriptstyle {\cal N}
\\
\scriptstyle w_2 \scriptstyle \in \scriptstyle t_2
\end{array}
}
\frac
{\mbox{count($w_1, w_2\!$)}}
{\mbox{ambig($w_1\!$)} \mbox{ambig($w_2\!$)}}
\end{math}

Here $\mbox{ambig}(w)$ is the number of categories in
which $w$ appears.  It has the effect of dividing the
evidence from a training instance across all possible
categories for the words.  The normaliser ensures
that all parameters for a head noun sum to unity.  

\subsection{Analysing the Test Set}
Given the high level descriptions in section
\ref{intro_analysis} it remains only to formalise 
the decision process used to analyse a noun compound.  
Each test compound presents a set of possible analyses and
the goal is to choose which analysis is most likely.  
For three word compounds
it suffices to
compute the ratio of two probabilities, that of a 
left-branching analysis and that of a right-branching one.
If this ratio is greater than unity, then the 
left-branching analysis is chosen.  When it is less than
unity, a right-branching analysis is
chosen.\footnote{If either probability estimate is
zero, the other analysis is chosen.  If both are zero
the analysis is made as if the ratio were exactly
unity.}  If the ratio is exactly unity, the analyser
guesses left-branching, although this is fairly rare
for conceptual association as shown by the
experimental results below.

For the adjacency model, when the given compound
is $w_1 w_2 w_3$, we can estimate this ratio as:

\begin{equation} \label{adj_ratio}
R_{adj}
\,\, = \,\,
\frac
{
\sum_{
	\scriptstyle t_i \scriptstyle \in
	   \mbox{\scriptsize cats($\scriptstyle
w_i\!$)}
	}
\Pr(t_1 \rightarrow t_2)
}
{
\sum_{
	\scriptstyle t_i \scriptstyle \in
	   \mbox{\scriptsize cats($\scriptstyle
w_i\!$)}
	}
\Pr(t_2 \rightarrow t_3)
}
\end{equation}

For the dependency model, the ratio is:

\begin{equation} \label{dep_ratio}
R_{dep}
\,\, = \,\,
\frac
{
\sum_{
	\scriptstyle t_i \scriptstyle \in
	   \mbox{\scriptsize cats($\scriptstyle
w_i\!$)}
	}
\Pr(t_1 \rightarrow t_2) \Pr(t_2 \rightarrow t_3)
}
{
\sum_{
	\scriptstyle t_i \scriptstyle \in
	   \mbox{\scriptsize cats($\scriptstyle
w_i\!$)}
	}
\Pr(t_1 \rightarrow t_3) \Pr(t_2 \rightarrow t_3)
}
\end{equation}
In both cases, we sum over all possible categories
for the words in the compound.  Because the dependency
model equations have two
factors, they are affected more severely by
data sparseness.  If the probability estimate for
$\Pr(t_2 \rightarrow t_3)$ is zero 
for all possible categories $t_2$ and $t_3$ 
then both the numerator and the denominator will 
be zero.  This will conceal any preference given 
by the parameters involving $t_1$.  
In such cases, we observe that the
test instance itself provides the information that the
event $t_2 \rightarrow t_3$ can occur and we
recalculate the ratio using $\Pr(t_2 \rightarrow t_3)
= k$ for all possible categories $t_2, t_3$ where $k$
is any non-zero constant.  However, no correction is
made to the probability estimates for $\Pr(t_1
\rightarrow t_2)$ and $\Pr(t_1 \rightarrow t_3)$ for
unseen cases, thus putting the dependency model on
an equal footing with the adjacency model above.

The equations presented above for the dependency
model differ from those developed in Lauer and
Dras (1994) in one way.  There, an additional
weighting factor (of 2.0) is used to favour a left-branching
analysis.  This arises because their construction
is based on the dependency model which
predicts that left-branching analyses should occur
twice as often.  Also, the work reported in Lauer and
Dras (1994) uses simplistic estimates of the
probability of a word given its thesaurus category.
The equations above assume these probabilities are uniformly
constant.  Section \ref{results_tuning} below
shows the result of making these two additions to
the method.

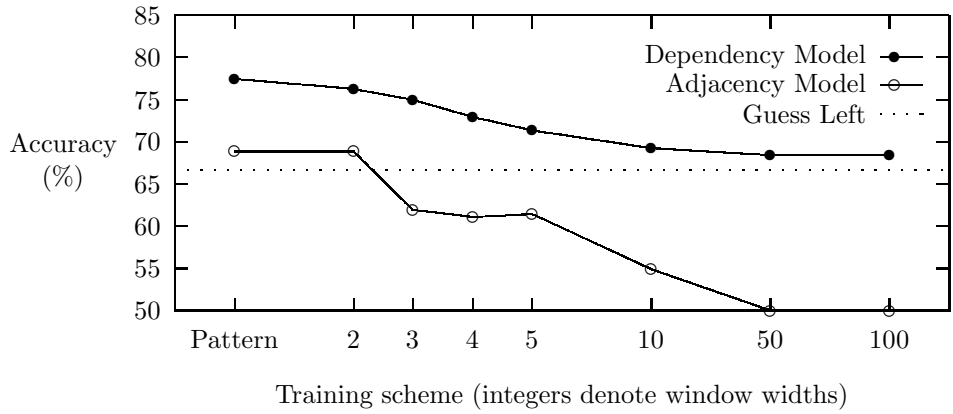
\begin{figure*}
\centering

\setlength{\unitlength}{0.240900pt}
\ifx\plotpoint\undefined\newsavebox{\plotpoint}\fi
\sbox{\plotpoint}{\rule[-0.200pt]{0.400pt}{0.400pt}}%
\begin{picture}(1500,600)(0,0)
\font\gnuplot=cmr10 at 10pt
\gnuplot
\sbox{\plotpoint}{\rule[-0.200pt]{0.400pt}{0.400pt}}%
\put(220.0,113.0){\rule[-0.200pt]{4.818pt}{0.400pt}}
\put(198,113){\makebox(0,0)[r]{50}}
\put(1416.0,113.0){\rule[-0.200pt]{4.818pt}{0.400pt}}
\put(220.0,179.0){\rule[-0.200pt]{4.818pt}{0.400pt}}
\put(198,179){\makebox(0,0)[r]{55}}
\put(1416.0,179.0){\rule[-0.200pt]{4.818pt}{0.400pt}}
\put(220.0,246.0){\rule[-0.200pt]{4.818pt}{0.400pt}}
\put(198,246){\makebox(0,0)[r]{60}}
\put(1416.0,246.0){\rule[-0.200pt]{4.818pt}{0.400pt}}
\put(220.0,312.0){\rule[-0.200pt]{4.818pt}{0.400pt}}
\put(198,312){\makebox(0,0)[r]{65}}
\put(1416.0,312.0){\rule[-0.200pt]{4.818pt}{0.400pt}}
\put(220.0,378.0){\rule[-0.200pt]{4.818pt}{0.400pt}}
\put(198,378){\makebox(0,0)[r]{70}}
\put(1416.0,378.0){\rule[-0.200pt]{4.818pt}{0.400pt}}
\put(220.0,444.0){\rule[-0.200pt]{4.818pt}{0.400pt}}
\put(198,444){\makebox(0,0)[r]{75}}
\put(1416.0,444.0){\rule[-0.200pt]{4.818pt}{0.400pt}}
\put(220.0,511.0){\rule[-0.200pt]{4.818pt}{0.400pt}}
\put(198,511){\makebox(0,0)[r]{80}}
\put(1416.0,511.0){\rule[-0.200pt]{4.818pt}{0.400pt}}
\put(220.0,577.0){\rule[-0.200pt]{4.818pt}{0.400pt}}
\put(198,577){\makebox(0,0)[r]{85}}
\put(1416.0,577.0){\rule[-0.200pt]{4.818pt}{0.400pt}}
\put(220.0,113.0){\rule[-0.200pt]{0.400pt}{4.818pt}}
\put(220.0,557.0){\rule[-0.200pt]{0.400pt}{4.818pt}}
\put(314.0,113.0){\rule[-0.200pt]{0.400pt}{4.818pt}}
\put(314,68){\makebox(0,0){Pattern}}
\put(314.0,557.0){\rule[-0.200pt]{0.400pt}{4.818pt}}
\put(501.0,113.0){\rule[-0.200pt]{0.400pt}{4.818pt}}
\put(501,68){\makebox(0,0){2}}
\put(501.0,557.0){\rule[-0.200pt]{0.400pt}{4.818pt}}
\put(594.0,113.0){\rule[-0.200pt]{0.400pt}{4.818pt}}
\put(594,68){\makebox(0,0){3}}
\put(594.0,557.0){\rule[-0.200pt]{0.400pt}{4.818pt}}
\put(688.0,113.0){\rule[-0.200pt]{0.400pt}{4.818pt}}
\put(688,68){\makebox(0,0){4}}
\put(688.0,557.0){\rule[-0.200pt]{0.400pt}{4.818pt}}
\put(781.0,113.0){\rule[-0.200pt]{0.400pt}{4.818pt}}
\put(781,68){\makebox(0,0){5}}
\put(781.0,557.0){\rule[-0.200pt]{0.400pt}{4.818pt}}
\put(968.0,113.0){\rule[-0.200pt]{0.400pt}{4.818pt}}
\put(968,68){\makebox(0,0){10}}
\put(968.0,557.0){\rule[-0.200pt]{0.400pt}{4.818pt}}
\put(1155.0,113.0){\rule[-0.200pt]{0.400pt}{4.818pt}}
\put(1155,68){\makebox(0,0){50}}
\put(1155.0,557.0){\rule[-0.200pt]{0.400pt}{4.818pt}}
\put(1342.0,113.0){\rule[-0.200pt]{0.400pt}{4.818pt}}
\put(1342,68){\makebox(0,0){100}}
\put(1342.0,557.0){\rule[-0.200pt]{0.400pt}{4.818pt}}
\put(220.0,113.0){\rule[-0.200pt]{292.934pt}{0.400pt}}
\put(1436.0,113.0){\rule[-0.200pt]{0.400pt}{111.778pt}}
\put(220.0,577.0){\rule[-0.200pt]{292.934pt}{0.400pt}}
\put(45,345){\makebox(0,0){\shortstack{Accuracy\\(\%)}}}
\put(828,-22){\makebox(0,0){Training scheme (integers denote window widths)}}
\put(220.0,113.0){\rule[-0.200pt]{0.400pt}{111.778pt}}
\put(1306,512){\makebox(0,0)[r]{Dependency Model}}
\put(1328.0,512.0){\rule[-0.200pt]{15.899pt}{0.400pt}}
\put(314,477){\usebox{\plotpoint}}
\multiput(314.00,475.92)(5.971,-0.494){29}{\rule{4.775pt}{0.119pt}}
\multiput(314.00,476.17)(177.089,-16.000){2}{\rule{2.387pt}{0.400pt}}
\multiput(501.00,459.92)(2.783,-0.495){31}{\rule{2.288pt}{0.119pt}}
\multiput(501.00,460.17)(88.251,-17.000){2}{\rule{1.144pt}{0.400pt}}
\multiput(594.00,442.92)(1.756,-0.497){51}{\rule{1.493pt}{0.120pt}}
\multiput(594.00,443.17)(90.902,-27.000){2}{\rule{0.746pt}{0.400pt}}
\multiput(688.00,415.92)(2.243,-0.496){39}{\rule{1.871pt}{0.119pt}}
\multiput(688.00,416.17)(89.116,-21.000){2}{\rule{0.936pt}{0.400pt}}
\multiput(781.00,394.92)(3.375,-0.497){53}{\rule{2.771pt}{0.120pt}}
\multiput(781.00,395.17)(181.248,-28.000){2}{\rule{1.386pt}{0.400pt}}
\multiput(968.00,366.92)(8.798,-0.492){19}{\rule{6.900pt}{0.118pt}}
\multiput(968.00,367.17)(172.679,-11.000){2}{\rule{3.450pt}{0.400pt}}
\put(1350,512){\circle*{18}}
\put(314,477){\circle*{18}}
\put(501,461){\circle*{18}}
\put(594,444){\circle*{18}}
\put(688,417){\circle*{18}}
\put(781,396){\circle*{18}}
\put(968,368){\circle*{18}}
\put(1155,357){\circle*{18}}
\put(1342,357){\circle*{18}}
\put(1155.0,357.0){\rule[-0.200pt]{45.048pt}{0.400pt}}
\put(1306,467){\makebox(0,0)[r]{Adjacency Model}}
\put(1328.0,467.0){\rule[-0.200pt]{15.899pt}{0.400pt}}
\put(314,363){\usebox{\plotpoint}}
\multiput(501.00,361.92)(0.505,-0.499){181}{\rule{0.504pt}{0.120pt}}
\multiput(501.00,362.17)(91.953,-92.000){2}{\rule{0.252pt}{0.400pt}}
\multiput(594.00,269.92)(4.411,-0.492){19}{\rule{3.518pt}{0.118pt}}
\multiput(594.00,270.17)(86.698,-11.000){2}{\rule{1.759pt}{0.400pt}}
\multiput(688.00,260.59)(10.283,0.477){7}{\rule{7.540pt}{0.115pt}}
\multiput(688.00,259.17)(77.350,5.000){2}{\rule{3.770pt}{0.400pt}}
\multiput(781.00,263.92)(1.077,-0.499){171}{\rule{0.960pt}{0.120pt}}
\multiput(781.00,264.17)(185.008,-87.000){2}{\rule{0.480pt}{0.400pt}}
\multiput(968.00,176.92)(1.443,-0.499){127}{\rule{1.251pt}{0.120pt}}
\multiput(968.00,177.17)(184.404,-65.000){2}{\rule{0.625pt}{0.400pt}}
\put(314.0,363.0){\rule[-0.200pt]{45.048pt}{0.400pt}}
\put(1350,467){\circle{18}}
\put(314,363){\circle{18}}
\put(501,363){\circle{18}}
\put(594,271){\circle{18}}
\put(688,260){\circle{18}}
\put(781,265){\circle{18}}
\put(968,178){\circle{18}}
\put(1155,113){\circle{18}}
\put(1342,113){\circle{18}}
\put(1155.0,113.0){\rule[-0.200pt]{45.048pt}{0.400pt}}
\put(1306,422){\makebox(0,0)[r]{Guess Left}}
\multiput(1328,422)(20.756,0.000){4}{\usebox{\plotpoint}}
\put(1394,422){\usebox{\plotpoint}}
\put(220,334){\usebox{\plotpoint}}
\put(220.00,334.00){\usebox{\plotpoint}}
\put(240.76,334.00){\usebox{\plotpoint}}
\multiput(245,334)(20.756,0.000){0}{\usebox{\plotpoint}}
\put(261.51,334.00){\usebox{\plotpoint}}
\multiput(269,334)(20.756,0.000){0}{\usebox{\plotpoint}}
\put(282.27,334.00){\usebox{\plotpoint}}
\put(303.02,334.00){\usebox{\plotpoint}}
\multiput(306,334)(20.756,0.000){0}{\usebox{\plotpoint}}
\put(323.78,334.00){\usebox{\plotpoint}}
\multiput(331,334)(20.756,0.000){0}{\usebox{\plotpoint}}
\put(344.53,334.00){\usebox{\plotpoint}}
\put(365.29,334.00){\usebox{\plotpoint}}
\multiput(367,334)(20.756,0.000){0}{\usebox{\plotpoint}}
\put(386.04,334.00){\usebox{\plotpoint}}
\multiput(392,334)(20.756,0.000){0}{\usebox{\plotpoint}}
\put(406.80,334.00){\usebox{\plotpoint}}
\put(427.55,334.00){\usebox{\plotpoint}}
\multiput(429,334)(20.756,0.000){0}{\usebox{\plotpoint}}
\put(448.31,334.00){\usebox{\plotpoint}}
\multiput(453,334)(20.756,0.000){0}{\usebox{\plotpoint}}
\put(469.07,334.00){\usebox{\plotpoint}}
\put(489.82,334.00){\usebox{\plotpoint}}
\multiput(490,334)(20.756,0.000){0}{\usebox{\plotpoint}}
\put(510.58,334.00){\usebox{\plotpoint}}
\multiput(515,334)(20.756,0.000){0}{\usebox{\plotpoint}}
\put(531.33,334.00){\usebox{\plotpoint}}
\multiput(539,334)(20.756,0.000){0}{\usebox{\plotpoint}}
\put(552.09,334.00){\usebox{\plotpoint}}
\put(572.84,334.00){\usebox{\plotpoint}}
\multiput(576,334)(20.756,0.000){0}{\usebox{\plotpoint}}
\put(593.60,334.00){\usebox{\plotpoint}}
\multiput(601,334)(20.756,0.000){0}{\usebox{\plotpoint}}
\put(614.35,334.00){\usebox{\plotpoint}}
\put(635.11,334.00){\usebox{\plotpoint}}
\multiput(638,334)(20.756,0.000){0}{\usebox{\plotpoint}}
\put(655.87,334.00){\usebox{\plotpoint}}
\multiput(662,334)(20.756,0.000){0}{\usebox{\plotpoint}}
\put(676.62,334.00){\usebox{\plotpoint}}
\put(697.38,334.00){\usebox{\plotpoint}}
\multiput(699,334)(20.756,0.000){0}{\usebox{\plotpoint}}
\put(718.13,334.00){\usebox{\plotpoint}}
\multiput(724,334)(20.756,0.000){0}{\usebox{\plotpoint}}
\put(738.89,334.00){\usebox{\plotpoint}}
\put(759.64,334.00){\usebox{\plotpoint}}
\multiput(760,334)(20.756,0.000){0}{\usebox{\plotpoint}}
\put(780.40,334.00){\usebox{\plotpoint}}
\multiput(785,334)(20.756,0.000){0}{\usebox{\plotpoint}}
\put(801.15,334.00){\usebox{\plotpoint}}
\put(821.91,334.00){\usebox{\plotpoint}}
\multiput(822,334)(20.756,0.000){0}{\usebox{\plotpoint}}
\put(842.66,334.00){\usebox{\plotpoint}}
\multiput(846,334)(20.756,0.000){0}{\usebox{\plotpoint}}
\put(863.42,334.00){\usebox{\plotpoint}}
\multiput(871,334)(20.756,0.000){0}{\usebox{\plotpoint}}
\put(884.18,334.00){\usebox{\plotpoint}}
\put(904.93,334.00){\usebox{\plotpoint}}
\multiput(908,334)(20.756,0.000){0}{\usebox{\plotpoint}}
\put(925.69,334.00){\usebox{\plotpoint}}
\multiput(932,334)(20.756,0.000){0}{\usebox{\plotpoint}}
\put(946.44,334.00){\usebox{\plotpoint}}
\put(967.20,334.00){\usebox{\plotpoint}}
\multiput(969,334)(20.756,0.000){0}{\usebox{\plotpoint}}
\put(987.95,334.00){\usebox{\plotpoint}}
\multiput(994,334)(20.756,0.000){0}{\usebox{\plotpoint}}
\put(1008.71,334.00){\usebox{\plotpoint}}
\put(1029.46,334.00){\usebox{\plotpoint}}
\multiput(1031,334)(20.756,0.000){0}{\usebox{\plotpoint}}
\put(1050.22,334.00){\usebox{\plotpoint}}
\multiput(1055,334)(20.756,0.000){0}{\usebox{\plotpoint}}
\put(1070.98,334.00){\usebox{\plotpoint}}
\put(1091.73,334.00){\usebox{\plotpoint}}
\multiput(1092,334)(20.756,0.000){0}{\usebox{\plotpoint}}
\put(1112.49,334.00){\usebox{\plotpoint}}
\multiput(1117,334)(20.756,0.000){0}{\usebox{\plotpoint}}
\put(1133.24,334.00){\usebox{\plotpoint}}
\multiput(1141,334)(20.756,0.000){0}{\usebox{\plotpoint}}
\put(1154.00,334.00){\usebox{\plotpoint}}
\put(1174.75,334.00){\usebox{\plotpoint}}
\multiput(1178,334)(20.756,0.000){0}{\usebox{\plotpoint}}
\put(1195.51,334.00){\usebox{\plotpoint}}
\multiput(1203,334)(20.756,0.000){0}{\usebox{\plotpoint}}
\put(1216.26,334.00){\usebox{\plotpoint}}
\put(1237.02,334.00){\usebox{\plotpoint}}
\multiput(1239,334)(20.756,0.000){0}{\usebox{\plotpoint}}
\put(1257.77,334.00){\usebox{\plotpoint}}
\multiput(1264,334)(20.756,0.000){0}{\usebox{\plotpoint}}
\put(1278.53,334.00){\usebox{\plotpoint}}
\put(1299.29,334.00){\usebox{\plotpoint}}
\multiput(1301,334)(20.756,0.000){0}{\usebox{\plotpoint}}
\put(1320.04,334.00){\usebox{\plotpoint}}
\multiput(1325,334)(20.756,0.000){0}{\usebox{\plotpoint}}
\put(1340.80,334.00){\usebox{\plotpoint}}
\put(1361.55,334.00){\usebox{\plotpoint}}
\multiput(1362,334)(20.756,0.000){0}{\usebox{\plotpoint}}
\put(1382.31,334.00){\usebox{\plotpoint}}
\multiput(1387,334)(20.756,0.000){0}{\usebox{\plotpoint}}
\put(1403.06,334.00){\usebox{\plotpoint}}
\put(1423.82,334.00){\usebox{\plotpoint}}
\multiput(1424,334)(20.756,0.000){0}{\usebox{\plotpoint}}
\put(1436,334){\usebox{\plotpoint}}
\end{picture}

\caption{Accuracy of dependency and adjacency model for
various training schemes} \label{dva_accuracy}
\end{figure*}

\section{Results}

\subsection{Dependency meets Adjacency}
Eight different training schemes have been used to
estimate the parameters and each set of estimates
used to analyse the test set under both the adjacency
and the dependency model.  The schemes used are:
\begin{itemize}
\item the pattern given in section \ref{method_train}; and
\item windowed training schemes with
window widths of 2, 3, 4, 5, 10, 50 and 100 words.
\end{itemize}

The accuracy on the test set for all these
experiments is shown in figure \ref{dva_accuracy}.
As can be seen, the dependency model is more accurate than the 
adjacency model.  This is true across the
whole spectrum of training schemes.
The proportion of cases in which the procedure was forced to 
guess, either because no data supported either 
analysis or because both were equally supported, is 
quite low.  For the pattern and two-word window training
schemes, the guess rate is less than 4\% for both models.  
In the three-word window training scheme, the guess
rates are less than 1\%.  For all larger windows,
neither model is ever forced to guess.

In the case of the pattern training scheme, the
difference between 68.9\% for adjacency and 77.5\%
for dependency is statistically significant at the 5\%
level ($p = 0.0316$), demonstrating the superiority
of the dependency model, at least for the compounds 
within Grolier's encyclopedia.

In no case do any of the windowed training schemes outperform 
the pattern scheme.  It seems that additional instances admitted
by the windowed schemes are too noisy to make an improvement.

Initial results from applying these methods to the
{\sc ema} corpus have been obtained by Wilco ter Stal 
(1995), and support the conclusion 
that the dependency model is
superior to the adjacency model.

\subsection{Tuning}  \label{results_tuning}
\begin{figure*}
\centering

\setlength{\unitlength}{0.240900pt}
\ifx\plotpoint\undefined\newsavebox{\plotpoint}\fi
\begin{picture}(1500,600)(0,0)
\font\gnuplot=cmr10 at 10pt
\gnuplot
\sbox{\plotpoint}{\rule[-0.200pt]{0.400pt}{0.400pt}}%
\put(220.0,113.0){\rule[-0.200pt]{4.818pt}{0.400pt}}
\put(198,113){\makebox(0,0)[r]{50}}
\put(1416.0,113.0){\rule[-0.200pt]{4.818pt}{0.400pt}}
\put(220.0,179.0){\rule[-0.200pt]{4.818pt}{0.400pt}}
\put(198,179){\makebox(0,0)[r]{55}}
\put(1416.0,179.0){\rule[-0.200pt]{4.818pt}{0.400pt}}
\put(220.0,246.0){\rule[-0.200pt]{4.818pt}{0.400pt}}
\put(198,246){\makebox(0,0)[r]{60}}
\put(1416.0,246.0){\rule[-0.200pt]{4.818pt}{0.400pt}}
\put(220.0,312.0){\rule[-0.200pt]{4.818pt}{0.400pt}}
\put(198,312){\makebox(0,0)[r]{65}}
\put(1416.0,312.0){\rule[-0.200pt]{4.818pt}{0.400pt}}
\put(220.0,378.0){\rule[-0.200pt]{4.818pt}{0.400pt}}
\put(198,378){\makebox(0,0)[r]{70}}
\put(1416.0,378.0){\rule[-0.200pt]{4.818pt}{0.400pt}}
\put(220.0,444.0){\rule[-0.200pt]{4.818pt}{0.400pt}}
\put(198,444){\makebox(0,0)[r]{75}}
\put(1416.0,444.0){\rule[-0.200pt]{4.818pt}{0.400pt}}
\put(220.0,511.0){\rule[-0.200pt]{4.818pt}{0.400pt}}
\put(198,511){\makebox(0,0)[r]{80}}
\put(1416.0,511.0){\rule[-0.200pt]{4.818pt}{0.400pt}}
\put(220.0,577.0){\rule[-0.200pt]{4.818pt}{0.400pt}}
\put(198,577){\makebox(0,0)[r]{85}}
\put(1416.0,577.0){\rule[-0.200pt]{4.818pt}{0.400pt}}
\put(220.0,113.0){\rule[-0.200pt]{0.400pt}{4.818pt}}
\put(220.0,557.0){\rule[-0.200pt]{0.400pt}{4.818pt}}
\put(372.0,113.0){\rule[-0.200pt]{0.400pt}{4.818pt}}
\put(372,68){\makebox(0,0){Pattern}}
\put(372.0,557.0){\rule[-0.200pt]{0.400pt}{4.818pt}}
\put(676.0,113.0){\rule[-0.200pt]{0.400pt}{4.818pt}}
\put(676,68){\makebox(0,0){2}}
\put(676.0,557.0){\rule[-0.200pt]{0.400pt}{4.818pt}}
\put(828.0,113.0){\rule[-0.200pt]{0.400pt}{4.818pt}}
\put(828,68){\makebox(0,0){3}}
\put(828.0,557.0){\rule[-0.200pt]{0.400pt}{4.818pt}}
\put(1132.0,113.0){\rule[-0.200pt]{0.400pt}{4.818pt}}
\put(1132,68){\makebox(0,0){5}}
\put(1132.0,557.0){\rule[-0.200pt]{0.400pt}{4.818pt}}
\put(1436.0,113.0){\rule[-0.200pt]{0.400pt}{4.818pt}}
\put(1436,68){\makebox(0,0){10}}
\put(1436.0,557.0){\rule[-0.200pt]{0.400pt}{4.818pt}}
\put(220.0,113.0){\rule[-0.200pt]{292.934pt}{0.400pt}}
\put(1436.0,113.0){\rule[-0.200pt]{0.400pt}{111.778pt}}
\put(220.0,577.0){\rule[-0.200pt]{292.934pt}{0.400pt}}
\put(45,345){\makebox(0,0){\shortstack{Accuracy\\(\%)}}}
\put(828,-22){\makebox(0,0){Training scheme (integers denote window widths)}}
\put(220.0,113.0){\rule[-0.200pt]{0.400pt}{111.778pt}}
\put(1306,512){\makebox(0,0)[r]{Tuned Dependency}}
\put(1328.0,512.0){\rule[-0.200pt]{15.899pt}{0.400pt}}
\put(372,488){\usebox{\plotpoint}}
\multiput(676.00,486.92)(1.558,-0.498){95}{\rule{1.341pt}{0.120pt}}
\multiput(676.00,487.17)(149.217,-49.000){2}{\rule{0.670pt}{0.400pt}}
\multiput(828.00,437.92)(9.716,-0.494){29}{\rule{7.700pt}{0.119pt}}
\multiput(828.00,438.17)(288.018,-16.000){2}{\rule{3.850pt}{0.400pt}}
\multiput(1132.00,421.93)(27.432,-0.482){9}{\rule{20.367pt}{0.116pt}}
\multiput(1132.00,422.17)(261.728,-6.000){2}{\rule{10.183pt}{0.400pt}}
\put(1350,512){\circle*{18}}
\put(372,488){\circle*{18}}
\put(676,488){\circle*{18}}
\put(828,439){\circle*{18}}
\put(1132,423){\circle*{18}}
\put(1436,417){\circle*{18}}
\put(372.0,488.0){\rule[-0.200pt]{73.234pt}{0.400pt}}
\put(372,477){\usebox{\plotpoint}}
\multiput(372,477)(20.727,-1.091){15}{\usebox{\plotpoint}}
\multiput(676,461)(20.627,-2.307){8}{\usebox{\plotpoint}}
\multiput(828,444)(20.502,-3.237){14}{\usebox{\plotpoint}}
\multiput(1132,396)(20.668,-1.904){15}{\usebox{\plotpoint}}
\put(1436,368){\usebox{\plotpoint}}
\put(1306,467){\makebox(0,0)[r]{Tuned Adjacency}}
\put(1328.0,467.0){\rule[-0.200pt]{15.899pt}{0.400pt}}
\put(372,423){\usebox{\plotpoint}}
\multiput(676.00,421.92)(1.736,-0.498){85}{\rule{1.482pt}{0.120pt}}
\multiput(676.00,422.17)(148.924,-44.000){2}{\rule{0.741pt}{0.400pt}}
\multiput(828.00,377.92)(2.545,-0.499){117}{\rule{2.127pt}{0.120pt}}
\multiput(828.00,378.17)(299.586,-60.000){2}{\rule{1.063pt}{0.400pt}}
\multiput(1132.00,317.92)(3.559,-0.498){83}{\rule{2.928pt}{0.120pt}}
\multiput(1132.00,318.17)(297.923,-43.000){2}{\rule{1.464pt}{0.400pt}}
\put(1350,467){\circle{18}}
\put(372,423){\circle{18}}
\put(676,423){\circle{18}}
\put(828,379){\circle{18}}
\put(1132,319){\circle{18}}
\put(1436,276){\circle{18}}
\put(372.0,423.0){\rule[-0.200pt]{73.234pt}{0.400pt}}
\put(372,363){\usebox{\plotpoint}}
\multiput(372,363)(20.756,0.000){15}{\usebox{\plotpoint}}
\multiput(676,363)(17.756,-10.747){9}{\usebox{\plotpoint}}
\multiput(828,271)(20.751,-0.410){14}{\usebox{\plotpoint}}
\multiput(1132,265)(19.954,-5.711){16}{\usebox{\plotpoint}}
\put(1436,178){\usebox{\plotpoint}}
\end{picture}

\caption{Accuracy of tuned dependency and adjacency model
for various training schemes} \label{tuned_accuracy}
\end{figure*}
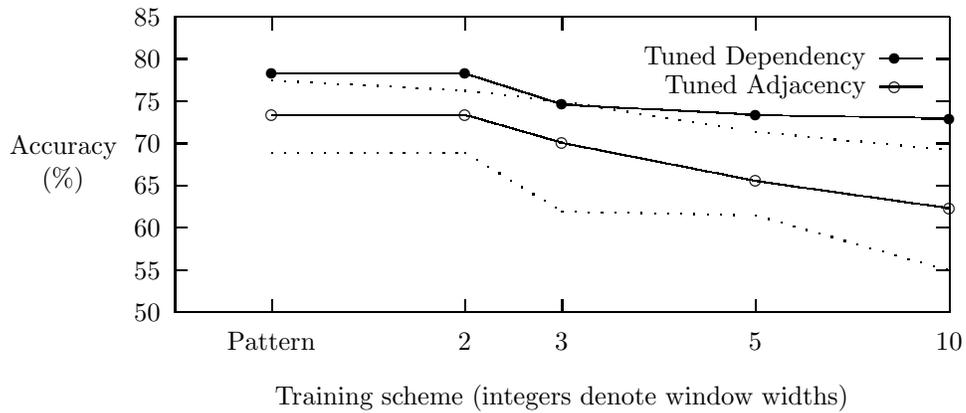

Lauer and Dras (1994) suggest two improvements to
the method used above.  These are:
\begin{itemize}
\item a factor favouring left-branching which arises
from the formal dependency construction; and
\item factors allowing for naive estimates of the
variation in the probability of categories.
\end{itemize}
While these changes are motivated by the dependency 
model, I have also applied them to the adjacency model
for comparison.
To implement them, equations
\ref{adj_ratio} and \ref{dep_ratio} must be
modified to incorporate a factor of $\frac{1}{\mid
t_1 \mid \mid t_2 \mid \mid t_3 \mid}$ in each
term of the sum and the entire ratio must be
multiplied by two.  Five training schemes 
have been applied with these extensions.  
The accuracy results are
shown in figure~\ref{tuned_accuracy}.  For
comparison, the untuned accuracy figures are shown
with dotted lines.  A marked improvement is
observed for the adjacency model, while the
dependency model is only slightly improved.

\subsection{Lexical Association} \label{results_lex}
To determine the difference made by conceptual 
association, the pattern training scheme has been retrained
using lexical counts for both the dependency
and adjacency model, but only for the words in the test set.
If the same system were to be applied across all of 
$\cal N$ (a total of 90,000 nouns), then around 8.1
billion parameters would be required.  
Left-branching is favoured by a factor of two
as described in the previous section, but no
estimates for the category probabilities are used
(these being meaningless for the lexical association
method).

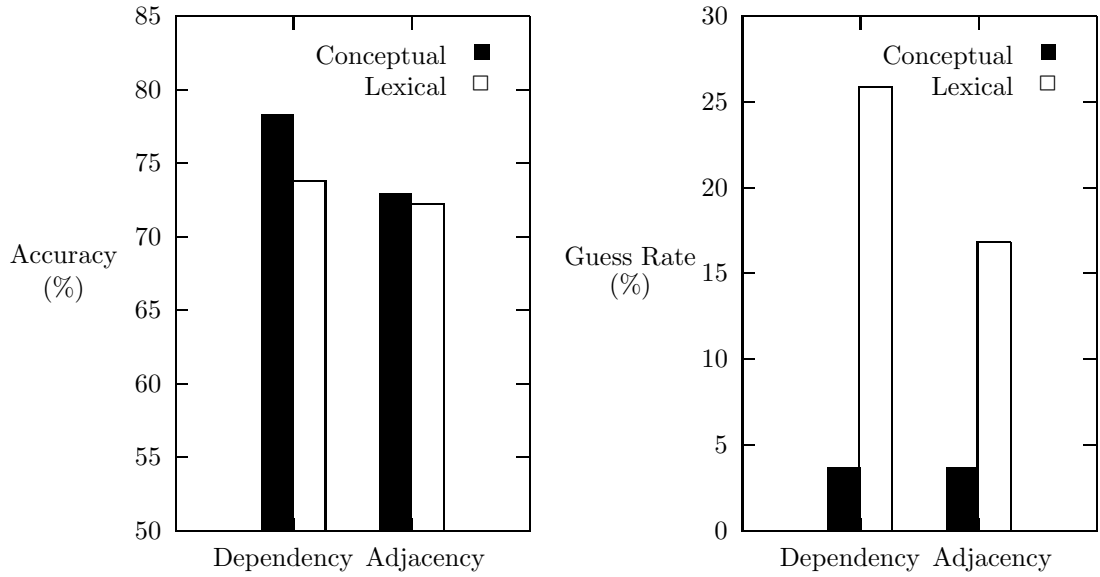
\begin{figure*}
\centering

\setlength{\unitlength}{0.240900pt}
\ifx\plotpoint\undefined\newsavebox{\plotpoint}\fi
\begin{picture}(1730, 900)(0,0)
\put(0,0){
\begin{picture}(840,900)(0,0)
\font\gnuplot=cmr10 at 10pt
\gnuplot
\sbox{\plotpoint}{\rule[-0.200pt]{0.400pt}{0.400pt}}%
\put(220.0,68.0){\rule[-0.200pt]{4.818pt}{0.400pt}}
\put(198,68){\makebox(0,0)[r]{50}}
\put(756.0,68.0){\rule[-0.200pt]{4.818pt}{0.400pt}}
\put(220.0,184.0){\rule[-0.200pt]{4.818pt}{0.400pt}}
\put(198,184){\makebox(0,0)[r]{55}}
\put(756.0,184.0){\rule[-0.200pt]{4.818pt}{0.400pt}}
\put(220.0,299.0){\rule[-0.200pt]{4.818pt}{0.400pt}}
\put(198,299){\makebox(0,0)[r]{60}}
\put(756.0,299.0){\rule[-0.200pt]{4.818pt}{0.400pt}}
\put(220.0,415.0){\rule[-0.200pt]{4.818pt}{0.400pt}}
\put(198,415){\makebox(0,0)[r]{65}}
\put(756.0,415.0){\rule[-0.200pt]{4.818pt}{0.400pt}}
\put(220.0,530.0){\rule[-0.200pt]{4.818pt}{0.400pt}}
\put(198,530){\makebox(0,0)[r]{70}}
\put(756.0,530.0){\rule[-0.200pt]{4.818pt}{0.400pt}}
\put(220.0,646.0){\rule[-0.200pt]{4.818pt}{0.400pt}}
\put(198,646){\makebox(0,0)[r]{75}}
\put(756.0,646.0){\rule[-0.200pt]{4.818pt}{0.400pt}}
\put(220.0,761.0){\rule[-0.200pt]{4.818pt}{0.400pt}}
\put(198,761){\makebox(0,0)[r]{80}}
\put(756.0,761.0){\rule[-0.200pt]{4.818pt}{0.400pt}}
\put(220.0,877.0){\rule[-0.200pt]{4.818pt}{0.400pt}}
\put(198,877){\makebox(0,0)[r]{85}}
\put(756.0,877.0){\rule[-0.200pt]{4.818pt}{0.400pt}}
\put(220.0,68.0){\rule[-0.200pt]{0.400pt}{4.818pt}}
\put(220.0,857.0){\rule[-0.200pt]{0.400pt}{4.818pt}}
\put(405.0,68.0){\rule[-0.200pt]{0.400pt}{4.818pt}}
\put(405,23){\makebox(0,0){Dependency\hspace{0.1in}}}
\put(405.0,857.0){\rule[-0.200pt]{0.400pt}{4.818pt}}
\put(591.0,68.0){\rule[-0.200pt]{0.400pt}{4.818pt}}
\put(591,23){\makebox(0,0){\hspace{0.1in} Adjacency}}
\put(591.0,857.0){\rule[-0.200pt]{0.400pt}{4.818pt}}
\put(220.0,68.0){\rule[-0.200pt]{133.940pt}{0.400pt}}
\put(776.0,68.0){\rule[-0.200pt]{0.400pt}{194.888pt}}
\put(220.0,877.0){\rule[-0.200pt]{133.940pt}{0.400pt}}
\put(45,472){\makebox(0,0){\shortstack{Accuracy\\(\%)}}}
\put(220.0,68.0){\rule[-0.200pt]{0.400pt}{194.888pt}}

\put(355,68){\rule{12.05pt}{157.55pt}}
\put(405,68){\framebox(50,549){\hspace{0.1in}}}
\put(541,68){\rule{12.05pt}{127.68pt}}
\put(591,68){\framebox(50,512){\hspace{0.1in}}}

\put(646,812){\makebox(0,0)[r]{Conceptual}}
\put(690,812){\raisebox{-.8pt}{\rule{5.5pt}{5.5pt}}}
\put(646,767){\makebox(0,0)[r]{Lexical}}
\put(690,767){\makebox(0,0){\hspace{5pt}$\Box$}}

\end{picture}
}
\put(890,0){
\begin{picture}(840,900)(0,0)
\font\gnuplot=cmr10 at 10pt
\gnuplot
\sbox{\plotpoint}{\rule[-0.200pt]{0.400pt}{0.400pt}}%
\put(220.0,68.0){\rule[-0.200pt]{133.940pt}{0.400pt}}
\put(220.0,68.0){\rule[-0.200pt]{4.818pt}{0.400pt}}
\put(198,68){\makebox(0,0)[r]{0}}
\put(756.0,68.0){\rule[-0.200pt]{4.818pt}{0.400pt}}
\put(220.0,203.0){\rule[-0.200pt]{4.818pt}{0.400pt}}
\put(198,203){\makebox(0,0)[r]{5}}
\put(756.0,203.0){\rule[-0.200pt]{4.818pt}{0.400pt}}
\put(220.0,338.0){\rule[-0.200pt]{4.818pt}{0.400pt}}
\put(198,338){\makebox(0,0)[r]{10}}
\put(756.0,338.0){\rule[-0.200pt]{4.818pt}{0.400pt}}
\put(220.0,473.0){\rule[-0.200pt]{4.818pt}{0.400pt}}
\put(198,473){\makebox(0,0)[r]{15}}
\put(756.0,473.0){\rule[-0.200pt]{4.818pt}{0.400pt}}
\put(220.0,607.0){\rule[-0.200pt]{4.818pt}{0.400pt}}
\put(198,607){\makebox(0,0)[r]{20}}
\put(756.0,607.0){\rule[-0.200pt]{4.818pt}{0.400pt}}
\put(220.0,742.0){\rule[-0.200pt]{4.818pt}{0.400pt}}
\put(198,742){\makebox(0,0)[r]{25}}
\put(756.0,742.0){\rule[-0.200pt]{4.818pt}{0.400pt}}
\put(220.0,877.0){\rule[-0.200pt]{4.818pt}{0.400pt}}
\put(198,877){\makebox(0,0)[r]{30}}
\put(756.0,877.0){\rule[-0.200pt]{4.818pt}{0.400pt}}
\put(220.0,68.0){\rule[-0.200pt]{0.400pt}{4.818pt}}
\put(220.0,857.0){\rule[-0.200pt]{0.400pt}{4.818pt}}
\put(405.0,68.0){\rule[-0.200pt]{0.400pt}{4.818pt}}
\put(405,23){\makebox(0,0){Dependency\hspace{0.1in}}}
\put(405.0,857.0){\rule[-0.200pt]{0.400pt}{4.818pt}}
\put(591.0,68.0){\rule[-0.200pt]{0.400pt}{4.818pt}}
\put(591,23){\makebox(0,0){\hspace{0.1in} Adjacency}}
\put(591.0,857.0){\rule[-0.200pt]{0.400pt}{4.818pt}}
\put(220.0,68.0){\rule[-0.200pt]{133.940pt}{0.400pt}}
\put(776.0,68.0){\rule[-0.200pt]{0.400pt}{194.888pt}}
\put(220.0,877.0){\rule[-0.200pt]{133.940pt}{0.400pt}}
\put(45,472){\makebox(0,0){\shortstack{Guess Rate\\(\%)}}}
\put(220.0,68.0){\rule[-0.200pt]{0.400pt}{194.888pt}}

\put(355,68){\rule{12.05pt}{24.09pt}}
\put(405,68){\framebox(50,696){\hspace{0.1in}}}
\put(541,68){\rule{12.05pt}{24.09pt}}
\put(591,68){\framebox(50,453){\hspace{0.1in}}}

\put(646,812){\makebox(0,0)[r]{Conceptual}}
\put(690,812){\raisebox{-.8pt}{\rule{5.5pt}{5.5pt}}}
\put(646,767){\makebox(0,0)[r]{Lexical}}
\put(690,767){\makebox(0,0){\hspace{5pt}$\Box$}}
\end{picture}
}
\end{picture}

\caption{Accuracy and Guess Rates of Lexical and Conceptual Association}
\label{lex_accuracy}
\end{figure*}

Accuracy and guess rates are shown in figure
\ref{lex_accuracy}.  Conceptual
association outperforms lexical association,
presumably because of its ability to generalise.  

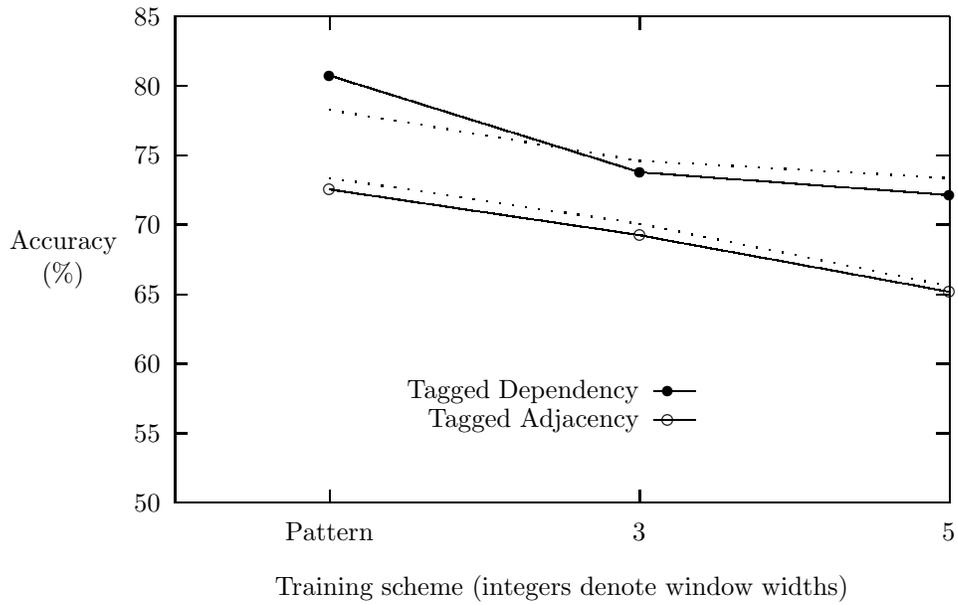
\begin{figure*}
\centering

\setlength{\unitlength}{0.240900pt}
\ifx\plotpoint\undefined\newsavebox{\plotpoint}\fi
\sbox{\plotpoint}{\rule[-0.200pt]{0.400pt}{0.400pt}}%
\begin{picture}(1500,900)(0,0)
\font\gnuplot=cmr10 at 10pt
\gnuplot
\sbox{\plotpoint}{\rule[-0.200pt]{0.400pt}{0.400pt}}%
\put(220.0,113.0){\rule[-0.200pt]{4.818pt}{0.400pt}}
\put(198,113){\makebox(0,0)[r]{50}}
\put(1416.0,113.0){\rule[-0.200pt]{4.818pt}{0.400pt}}
\put(220.0,222.0){\rule[-0.200pt]{4.818pt}{0.400pt}}
\put(198,222){\makebox(0,0)[r]{55}}
\put(1416.0,222.0){\rule[-0.200pt]{4.818pt}{0.400pt}}
\put(220.0,331.0){\rule[-0.200pt]{4.818pt}{0.400pt}}
\put(198,331){\makebox(0,0)[r]{60}}
\put(1416.0,331.0){\rule[-0.200pt]{4.818pt}{0.400pt}}
\put(220.0,440.0){\rule[-0.200pt]{4.818pt}{0.400pt}}
\put(198,440){\makebox(0,0)[r]{65}}
\put(1416.0,440.0){\rule[-0.200pt]{4.818pt}{0.400pt}}
\put(220.0,550.0){\rule[-0.200pt]{4.818pt}{0.400pt}}
\put(198,550){\makebox(0,0)[r]{70}}
\put(1416.0,550.0){\rule[-0.200pt]{4.818pt}{0.400pt}}
\put(220.0,659.0){\rule[-0.200pt]{4.818pt}{0.400pt}}
\put(198,659){\makebox(0,0)[r]{75}}
\put(1416.0,659.0){\rule[-0.200pt]{4.818pt}{0.400pt}}
\put(220.0,768.0){\rule[-0.200pt]{4.818pt}{0.400pt}}
\put(198,768){\makebox(0,0)[r]{80}}
\put(1416.0,768.0){\rule[-0.200pt]{4.818pt}{0.400pt}}
\put(220.0,877.0){\rule[-0.200pt]{4.818pt}{0.400pt}}
\put(198,877){\makebox(0,0)[r]{85}}
\put(1416.0,877.0){\rule[-0.200pt]{4.818pt}{0.400pt}}
\put(220.0,113.0){\rule[-0.200pt]{0.400pt}{4.818pt}}
\put(220.0,857.0){\rule[-0.200pt]{0.400pt}{4.818pt}}
\put(463.0,113.0){\rule[-0.200pt]{0.400pt}{4.818pt}}
\put(463,68){\makebox(0,0){Pattern}}
\put(463.0,857.0){\rule[-0.200pt]{0.400pt}{4.818pt}}
\put(950.0,113.0){\rule[-0.200pt]{0.400pt}{4.818pt}}
\put(950,68){\makebox(0,0){3}}
\put(950.0,857.0){\rule[-0.200pt]{0.400pt}{4.818pt}}
\put(1436.0,113.0){\rule[-0.200pt]{0.400pt}{4.818pt}}
\put(1436,68){\makebox(0,0){5}}
\put(1436.0,857.0){\rule[-0.200pt]{0.400pt}{4.818pt}}
\put(220.0,113.0){\rule[-0.200pt]{292.934pt}{0.400pt}}
\put(1436.0,113.0){\rule[-0.200pt]{0.400pt}{184.048pt}}
\put(220.0,877.0){\rule[-0.200pt]{292.934pt}{0.400pt}}
\put(45,495){\makebox(0,0){\shortstack{Accuracy\\(\%)}}}
\put(828,-22){\makebox(0,0){Training scheme (integers denote window widths)}}
\put(220.0,113.0){\rule[-0.200pt]{0.400pt}{184.048pt}}
\put(950,288){\makebox(0,0)[r]{Tagged Dependency}}
\put(972.0,288.0){\rule[-0.200pt]{15.899pt}{0.400pt}}
\put(463,784){\usebox{\plotpoint}}
\multiput(463.00,782.92)(1.604,-0.499){301}{\rule{1.382pt}{0.120pt}}
\multiput(463.00,783.17)(484.132,-152.000){2}{\rule{0.691pt}{0.400pt}}
\multiput(950.00,630.92)(6.811,-0.498){69}{\rule{5.500pt}{0.120pt}}
\multiput(950.00,631.17)(474.584,-36.000){2}{\rule{2.750pt}{0.400pt}}
\put(994,288){\circle*{18}}
\put(463,784){\circle*{18}}
\put(950,632){\circle*{18}}
\put(1436,596){\circle*{18}}
\put(463,730){\usebox{\plotpoint}}
\multiput(463,730)(20.481,-3.364){24}{\usebox{\plotpoint}}
\multiput(950,650)(20.724,-1.151){24}{\usebox{\plotpoint}}
\put(1436,623){\usebox{\plotpoint}}
\put(950,243){\makebox(0,0)[r]{Tagged Adjacency}}
\put(972.0,243.0){\rule[-0.200pt]{15.899pt}{0.400pt}}
\put(463,605){\usebox{\plotpoint}}
\multiput(463.00,603.92)(3.395,-0.499){141}{\rule{2.806pt}{0.120pt}}
\multiput(463.00,604.17)(481.177,-72.000){2}{\rule{1.403pt}{0.400pt}}
\multiput(950.00,531.92)(2.739,-0.499){175}{\rule{2.284pt}{0.120pt}}
\multiput(950.00,532.17)(481.259,-89.000){2}{\rule{1.142pt}{0.400pt}}
\put(994,243){\circle{18}}
\put(463,605){\circle{18}}
\put(950,533){\circle{18}}
\put(1436,444){\circle{18}}
\put(463,623){\usebox{\plotpoint}}
\multiput(463,623)(20.532,-3.036){24}{\usebox{\plotpoint}}
\multiput(950,551)(20.346,-4.103){24}{\usebox{\plotpoint}}
\put(1436,453){\usebox{\plotpoint}}
\end{picture}

\caption{Accuracy using a tagged corpus for various
training schemes}  \label{tag_accuracy}
\end{figure*}

\subsection{Using a Tagger}
One problem with the training methods given in
section \ref{method_train} is the restriction of
training data to nouns in $\cal N$.  Many nouns,
especially common ones, have verbal or adjectival
usages that preclude them from being in $\cal N$.
Yet when they occur as nouns, they still provide
useful training information that the current system
ignores.  To test whether using tagged data would
make a difference, the freely available Brill tagger
(Brill, 1993) was applied to the corpus.  Since no
manually tagged training data is available for our
corpus, the tagger's default rules were used (these
rules were produced by Brill by training on the 
Brown corpus).  This results in rather poor
tagging accuracy, so it is quite possible that a
manually tagged corpus would produce better
results.

Three training schemes have been used and the
tuned analysis procedures applied to the test set.
Figure \ref{tag_accuracy} shows the resulting
accuracy, with accuracy values from figure
\ref{tuned_accuracy} displayed with dotted lines.
If anything, admitting additional training data based
on the tagger introduces more noise, reducing the
accuracy.  However, for the pattern training scheme
an improvement was made to the dependency
model, producing the highest overall accuracy of 81\%.

\section{Conclusion}

The experiments above demonstrate a number of important points.
The most general of these is that even quite crude corpus 
statistics can provide information about the syntax of compound
nouns.  At the very least, this information can be applied in
broad coverage parsing to assist in the control of search.
I have also shown that with a corpus of moderate size it is
possible to get reasonable results without using a tagger
or parser by employing a customised training pattern.  While
using windowed co-occurrence did not help here, it is possible
that under more data sparse conditions better performance could
be achieved by this method.

The significance of the use of conceptual association deserves
some mention.  I have argued that without it a broad coverage system
would be impossible.  This is in contrast to previous work
on conceptual association where it resulted in little
improvement on a task which could already be performed.
In this study, not only has the technique proved its worth 
by supporting generality, but through generalisation of 
training information it outperforms the equivalent 
lexical association approach given the same information.

Amongst all the comparisons performed in these experiments
one stands out as exhibiting the greatest contrast.  
In all experiments the dependency model
provides a substantial advantage over the 
adjacency model, even though the latter is more prevalent
in proposals within the literature.  This result is in
accordance with the informal reasoning given in section
\ref{intro_analysis}.  The model also has the further
commendation that it predicts correctly the observed
proportion of left-branching compounds found in two
independently extracted test sets.

In all, the most accurate technique achieved an accuracy
of 81\% as compared to the 67\% achieved by 
guessing left-branching.
Given the high frequency of occurrence of noun
compounds in many texts, this suggests that the use of
these techniques in probabilistic parsers will result
in higher performance in broad coverage natural language
processing.

\section{Acknowledgements}

This work has received valuable input from people too numerous
to mention.  The most significant contributions have been
made by Richard Buckland, Robert Dale and Mark Dras.  I am
also indebted to Vance Gledhill, Mike Johnson,
Philip Resnik, Richard Sproat, Wilco ter~Stal,
Lucy Vanderwende and Wayne Wobcke.
Financial support is gratefully acknowledged from the Microsoft 
Institute and the Australian Government.

\end{document}